\documentclass[letterpaper, 10 pt, conference]{ieeeconf}  

\IEEEoverridecommandlockouts                              

\usepackage{graphics} 
\usepackage{epsfig} 
\usepackage{mathptmx} 
\usepackage{times} 
\usepackage{amsmath} 
\usepackage{amssymb}  
\usepackage{xspace}
\usepackage{xcolor}
\usepackage{multirow}
\usepackage{multicol}
\usepackage[utf8]{inputenc}
\usepackage{graphicx}
\usepackage{colortbl}
\usepackage{hyperref}

\usepackage{booktabs}

\definecolor{green}{RGB}{0,150,10}
\definecolor{blue}{RGB}{0,148,181}
\definecolor{orange}{RGB}{194,153,107}

\newcommand{\OURS}{\textsc{ADC}\xspace} %
\newcommand{\Ours}{\OURS} %

\title{Adversarial Data Collection: Human-Collaborative Perturbations for Efficient and Robust Robotic Imitation Learning}

\author{Siyuan Huang$^{*}$, Yue Liao$^{*}$, Siyuan Feng$^{*}$, Shu Jiang, Si Liu, Hongsheng Li, Maoqing Yao$^{\ddag}$, Guanghui Ren$^{\ddag}$
\thanks{$^{*}$ indicates the equal contribution and $^{\ddag}$ indicates corresponding authors.}%
\thanks{Siyuan Huang is with Shanghai Jiao Tong University. Yue Liao and Hongsheng Li are with MMLab, CUHK. Siyuan Feng, Shu Jiang, Maoqing Yao and Guanghui Ren are with Agibot. Si Liu is with Beihang University.}%
}

\begin{document}

\maketitle
\thispagestyle{empty}
\pagestyle{empty}


\begin{abstract}

The pursuit of data efficiency, where quality outweighs quantity, has emerged as a cornerstone in robotic manipulation, especially given the high costs associated with real-world data collection. We propose that maximizing the informational density of individual demonstrations can dramatically reduce reliance on large-scale datasets while improving task performance. To this end, we introduce \textbf{Adversarial Data Collection} (ADC), a Human-in-the-Loop (HiL) framework that redefines robotic data acquisition through real-time, bidirectional human-environment interactions. Unlike conventional pipelines that passively record static demonstrations, \Ours adopts a collaborative perturbation paradigm: during a single episode, an adversarial operator dynamically alters object states, environmental conditions, and linguistic commands, while the tele-operator adaptively adjusts actions to overcome these evolving challenges. This process compresses diverse failure-recovery behaviors, compositional task variations, and environmental perturbations into minimal demonstrations. Our experiments demonstrate that ADC-trained models achieve superior compositional generalization to unseen task instructions, enhanced robustness to perceptual perturbations, and emergent error recovery capabilities. Strikingly, models trained with merely 20\% of the demonstration volume collected through ADC significantly outperform traditional approaches using full datasets. These advances bridge the gap between data-centric learning paradigms and practical robotic deployment, demonstrating that strategic data acquisition, not merely post-hoc processing, is critical for scalable, real-world robot learning. 
 Additionally, we are curating a large-scale ADC-Robotics dataset comprising real-world manipulation tasks with adversarial perturbations. This benchmark will be open-sourced to facilitate advancements in robotic imitation learning. More information can be found on our project page: \href{https://sites.google.com/view/adc-robot}{https://sites.google.com/view/adc-robot}.

\end{abstract}

\section{Introduction}

In real-world robotic manipulation, collecting real-scene imitation learning data requires humans to teleoperate robotic arms to perform tasks based on linguistic instructions, which is a labor-intensive and costly process~\cite{AgiBot2024agibotworld, khazatsky2024droid}. Each demonstration demands significant human effort (\emph{e.g.}, expert operation, precise annotation), making it economically infeasible to scale data quantity for diverse environmental coverage. Real-world complexity—varying object configurations, lighting, and task constraints—further exacerbates the need for massive data diversity to ensure robust generalization~\cite{lin2024data}. While recent approaches attempt to address this by generating diversity through simulated environments~\cite{wang2023gensim,james2020rlbench} or low-cost Universal Manipulation Interface~(UMI) grippers~\cite{lin2024data,chi2024universal} (sampling one variation per trial, such as object poses), these methods face two fundamental limitations:
\begin{itemize}
    \item Synthetic data, despite its scalability, suffers from a domain gap due to idealized physics/kinematics, hindering transfer to real-robot dynamics;
    \item Extending such diversity to tele-operated systems remains impractical, as even minor environmental changes (\emph{e.g.}, repositioning objects) require costly human-in-the-loop resampling.
\end{itemize}

\begin{figure}[t]
    \centering
    \includegraphics[width=1\linewidth]{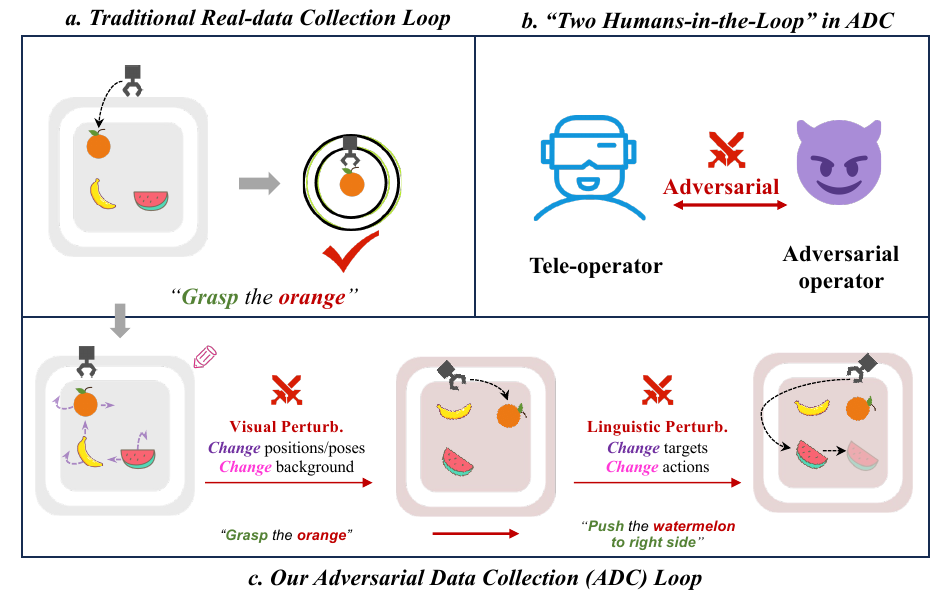}
    \caption{Comparative Analysis of the Real-Data Collection Loop in Robotic Manipulation. (a) Traditional Approach: A tele-operator executes tasks via fixed linguistic instructions in static visual environments.
    (b) Adversarial Data Collection~(ADC) Framework: Employs a Two-Humans-in-the-Loop approach, where a secondary operator intervenes to perturb the primary’s execution dynamically when the tele-operator is executing a task.
    (c) ADC Loop: The adversarial operator introduces visual (backgrounds, object positions/poses) and linguistic (task goals) perturbations, shifting environmental context and target objects within a single episode.
    }
    \label{fig:fig1}
\end{figure}

To contextualize our approach, we first analyze the data requirements and structural constraints of state-of-the-art robotic imitation frameworks, particularly Vision-Language-Action (VLA) models. VLAs integrate Multimodal Large Language Models (MLLMs) to unify reasoning over visual inputs, language instructions, and robotic actions. Most VLA~\cite{kim2024openvla, black2024pi_0} frameworks inherit the Markov assumption, predicting actions based solely on the current visual observation. This design necessitates training data in the form of triplets: current visual scene, language instruction, and action supervision. As shown in Fig.~\ref{fig:fig1}(a), Conventional data collection, however, treats an entire task (\emph{e.g.}, "grasping the orange") as a single demonstration, slicing it into sequential units. These units suffer from inherent inefficiencies—visually redundant frames, repetitive language instructions, and highly correlated actions—effectively diluting the informational value of each collected demonstration. For instance, a 30-second "pick-and-place" task might yield hundreds of near-identical training triplets, with minimal diversity in scenes or commands.

To overcome these limitations, we propose Adversarial Data Collection~(ADC), a collaborative Human-in-the-Loop~(HiL) framework that systematically injects controlled perturbations during data acquisition to maximize per-demonstration diversity and robustness. As illustrated in Fig.~\ref{fig:fig1}, the ADC framework operates through two synchronized human roles: a tele-operator executes primary manipulation tasks via standard robotic interfaces, while an adversarial operator dynamically modulates task parameters through real-time perturbations across perceptual, linguistic, and physical dimensions. During a single demonstration episode, the adversarial operator alters scene configurations (\emph{e.g.}, repositioning objects, rotating target objects mid-grasp and adjusting background) and paraphrases task instructions~(\emph{e.g.}, "grasping the orange" → "grasping the watermelon"). These perturbations force the tele-operator to adaptively re-plan actions, such as re-grasping strategies or trajectory corrections, under evolving constraints, thereby compressing recovery behaviors, instruction compositional generalization, and environmental variations into a unified demonstration. By systematically coupling human-driven adversity with real-time adaptation, \Ours not only elevates demonstration efficiency but also cultivates model robustness against perceptual ambiguities, linguistic variability, and physical uncertainties, which is critical for deploying VLAs in unstructured real-world environments

To evaluate the effectiveness of \OURS, we constructed a dataset consisting of $200$ demonstrations including balanced adversarially perturbed demonstrations ($96$K frames) from \Ours and demonstrations ($90$K frames). Each demonstration from \Ours incorporates multiple layers of real-time perturbations, spanning visual and linguistic dimensions, introduced dynamically during human tele-operation.  While marginally more time-intensive per demonstration, ADC-trained models achieve superior generalization and robustness with significantly fewer demonstrations—outperforming models trained on larger-scale conventional datasets in handling ambiguous instructions, environmental perturbations, and unseen object configurations. This efficiency stems from \OURS’s capacity to encode real-world complexity through intentional adversity, transforming data quality into a substitute for quantity—a critical advancement for scalable robotic learning. Additionally, we are curating a large-scale ADC-Robotics dataset comprising real-world manipulation tasks with adversarial perturbations. Upon completion, this benchmark will be open-sourced to facilitate advancements in robotic imitation learning.

\section{Related Work}
\begin{figure*}[ht]
    \centering

    \includegraphics[width=\linewidth]{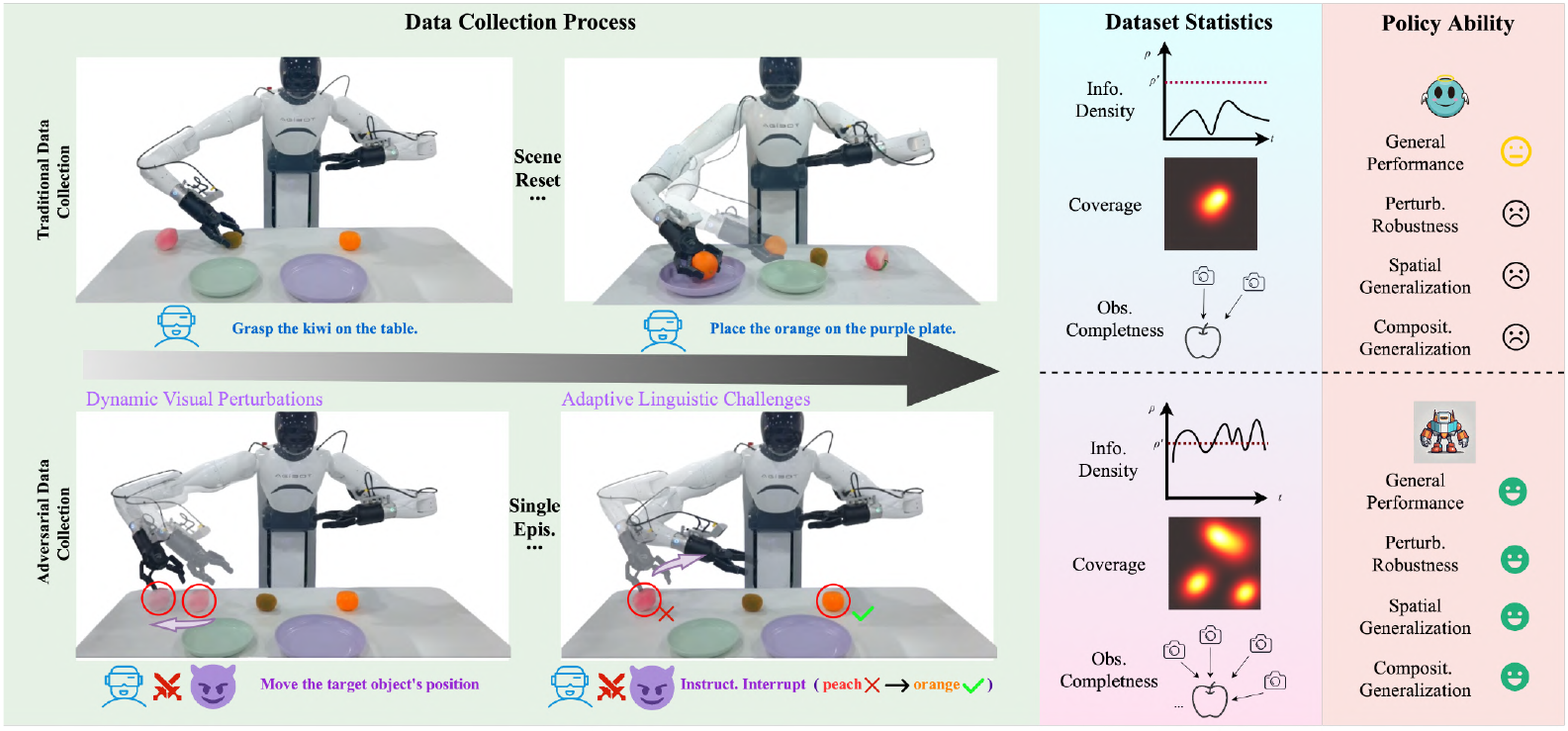}
    \caption{The overview of \Ours. During training data collection, we introduce several adversarial perturbations, including dynamic visual perturbations and adaptive linguistic challenges. These perturbations increase information density, expand state space coverage, and provide more complete observations of target objects. The resulting high-quality dataset enables the trained policy model to achieve strong robustness and generalization, outperforming models trained with conventional data collection strategies.}
    \label{fig:overview}
\end{figure*}
\subsection{Data Collection in Robot Learning} 
Data collection methods in robot learning have been extensively studied to improve generalization, generally with a focus on scaling up robot datasets~\cite{o2023open, walke2023bridgedata, khazatsky2024droid, AgiBot2024agibotworld}. These efforts have shown that greater object and task diversity can significantly improve policy robustness, highlighting the importance of collecting diverse datasets. For instance, Droid~\cite{khazatsky2024droid}, one of the largest existing datasets, contains 76k demonstration trajectories collected across more than 500 scenes. More recently, studies~\cite{gao2024efficient, lin2024data} have explored strategies for efficient data collection to enhance policy generalization; however, their discussions are largely constrained to compositional generalization with respect to visual factors. In this paper, we extend the scope of data collection by considering both visual and language dimensions, aiming to improve generalization across a broader range of factors. Additionally, we introduce real-time human intervention during the data collection process, dynamically perturbing robot actions rather than merely resetting initial conditions, thereby implicitly accounting for human-robot interaction in deployment scenarios.


\subsection{Vision-based Imitation Learning} 

Traditional imitation learning methods have relied on either state-based~\cite{ho2016generative} or image-based~\cite{young2021visual} representations to describe environments and goals. While state-based methods struggle to handle unstructured environments, image-based methods offer richer representations but face challenges related to goal ambiguity and specification. To address these issues, natural language has emerged as an intuitive and flexible way to specify goals, with recent work integrating language goals and image observations to enable more adaptable policy learning~\cite{shridhar2022cliport, jia2025x}. Additionally, fine-tuning VLMs into VLAs has further advanced this area~\cite{kim2024openvla, li2023generalist}. Furthermore, some studies have introduced diffusion models~\cite{song2021denoising} into policy learning to better model multi-modal action distributions~\cite{chi2023diffusion, liu2024rdt, hou2025diffusiontransformerpolicyscaling}. Meanwhile, other work has explored leveraging large-scale, unlabelled action videos to enhance the temporal and spatial understanding of policies~\cite{cheang2024gr, huang2025enerverse}. {As vision-based models scale up, they require increasingly diverse and high-quality robot data to avoid overfitting to shortcuts introduced by traditional data collection methods. To address this, we propose Adversarial Data Collection (ADC), which enhances information density through real-time perturbations.}

\subsection{Generalization and Robustness in Robotic Policy} Achieving both generalization and robustness in robotic policies is crucial, particularly in the context of dynamic human-robot interactions, where policies must not only generalize to unseen objects and environments but also remain robust to human interventions. Prior works have explored improving generalization through pre-trained visual representations~\cite{nair2022r3m} and diverse datasets~\cite{o2023open}, with some focusing on adapting to new environments~\cite{xing2021kitchenshift, xie2024decomposing} or objects~\cite{fang2023anygrasp, yu2024uniaff}. However, these efforts often overlook robustness in scenarios involving human intervention, which is critical for real-world deployment. Additionally, while compositional generalization—reasoning about unseen combinations of environmental factors—has been studied~\cite{xie2024decomposing, pumacay2024colosseum}, prior work primarily examines small-scale policies. As robotic learning transitions to large Vision-Language Action (VLA) models~\cite{kim2024openvla}, there is a growing need to enhance both generalization and robustness through fine-tuning on task-specific datasets. In this work, we address these challenges by systematically analyzing the interplay between generalization, robustness, and data collection, focusing on dynamic scenes involving human-robot interactions and scaling strategies for large VLA models.

\section{Approach}

To ground our approach, we first review the structure of training data in Vision-Language-Action (VLA) models (Sec.~\ref{sec:data_units}), then detail our Adversarial Data Collection (ADC) framework in Sec.~\ref{sec:adversarial}, and finally outline its integration into robotic imitation policies (Sec.~\ref{sec:c_policy}) and VLA training (Sec.~\ref{sec:method_in_vla}).

\subsection{VLA Data Units Density Analysis}
\label{sec:data_units}

\subsubsection{VLA Architecture and Training Data Structure}
The Visual-Language-Action (VLA) architecture adapts pre-trained multimodal Large Language Models~(MLLMs) for robotic control through residual policy heads. Focused on mainstream memory-less pipelines (\textit{e.g.}, OpenVLA~\cite{kim2024openvla}), these temporal-independent models process \textit{multi-view observations} ($\mathbb{V}_t^{\text{multi-view}}$) and \textit{language instructions} ($\mathbb{L}_t$) to predict actions $a_t$ via:
\begin{equation}
    p(a_t | \mathbb{V}_t^{\text{multi-view}}, \mathbb{L}_t) = \text{VLA}(\mathbb{V}_t, \mathbb{L}_t)
    \label{eq:markov}
\end{equation}
Training requires demonstration units $U_t \triangleq (\mathbb{V}_t, \mathbb{L}_t, a_t^*)$, where $a_t^*$ denotes expert actions. Traditional datasets assemble these units into fixed episodes $\mathbb{E}_i = \{U_1,...,U_n\}$ with static environments, causing two inefficiencies:  
\begin{itemize}
    \item \textbf{Intra-Episode Redundancy}: $>70$\% of consecutive $U_t$ share redundant visual-language contexts
    \item \textbf{Inter-Episode Fragmentation}: Functionally equivalent units remain isolated across episodes
\end{itemize}

\subsubsection{Density Optimization}
To overcome these limitations, we reformulate data quality via \textit{information density per demonstration}:
\begin{equation}
    \rho \triangleq \mathbb{E}[|\mathbb{U}(\mathbb{E})|],\quad \mathbb{U}(\mathbb{E}) \triangleq \{U_t | \nexists U_{t'} \prec U_t\}
    \label{eq:density}
\end{equation}
where $U_{t'} \prec U_t$ indicates functional equivalence under task constraints, including spatial positions, prompt diversity, etc. Instead of maximizing episode counts ($\max |\{\mathbb{E}_i\}|$), our adversarial strategy $\max_\text{perturb} \rho$ enriches unit diversity within individual demonstrations through:
\begin{itemize}
    \item \textbf{Multi-Order Variations}: Perturbing object poses, lighting, and instructions mid-episode
    \item \textbf{Compositional Coupling}: Synergistic combinations of visual-linguistic perturbations
\end{itemize}

As visualized in Fig.~\ref{fig:overview}, this approach compresses hundreds of traditional trials into single demonstrations by forcing adaptive responses to dynamic perturbations. The resultant dataset structure aligns with VLA's Markovian requirements—maximizing unique $U_t$ coverage while eliminating temporal redundancy.
\subsection{ADC: Adversarial HiL Framework}
\label{sec:adversarial}

ADC transforms conventional teleoperation into an adversarial HiL process, where two collaborative roles, the \textit{tele-operator} (executing tasks) and the \textit{adversarial operator} (introducing perturbations), interact dynamically during data collection. This framework introduces controlled perturbations across visual and linguistic dimensions to maximize per-demonstration diversity while preserving physical plausibility.

\noindent \subsubsection{Visual Perturbations}
The adversarial operator introduces physics-grounded visual disruptions through two key mechanisms:

\textbf{Positional Dynamics}: Target objects are initialized at positions sampled from task-specific Gaussian distributions $\mathcal{N}(\mu_{\text{task}}, \sigma^2_{\text{task}})$. If the end-effector (EEF) fails to enter the proximity threshold during grasping attempts, the adversarial operator randomly changes the objects' positions within the worktable area.

\textbf{Grasp Pose Disruptions}: Grasp Pose Disruptions:
When the end-effector enters a proximity threshold (set to 15 cm in practice), controlled perturbation impulses are applied to objects, altering their positions and orientations. This forces real-time grasp repositioning and reorientation, requiring the system to find the next-best grasp configuration.

These mechanisms are also applied to target containers, further enhancing task complexity.

 \subsubsection{Linguistic Perturbations}
Language instructions undergo contextual transformations to challenge static command assumptions:

 \textbf{Mid-Execution Interruption}: Task goals are dynamically altered during critical phases, including changes in the target object (e.g., "place cup" → "place bottle"), modifications to the required action (e.g., "pick up cup" → "push cup"), or simultaneous changes in both the object and the action (e.g., "pick up cup" → "push bottle"). These transformations require immediate adaptation while maintaining task stability.


 \textbf{Dynamic Spatial Redefinition}: Spatial descriptors such as "left" or "near" are continuously redefined relative to changing object positions, preventing overfitting to fixed coordinate systems.

In this paper, we primarily demonstrate the first mechanism, as spatial understanding for VLA is a separate research focus.

Through these coordinated perturbations, hundreds of static demonstration variations are compressed into unified episodes via real-time human adaptation, as illustrated in Fig.~\ref{fig:overview}. The tele-operator’s responsive behaviors—trajectory replanning, grasp recovery, and compositional task switching—generate rich and diverse data units $U_t = (\mathbb{V}_t^{\text{multi-view}}, \mathbb{L}_t, a_t^*)$ that conventional pipelines cannot achieve.
\begin{figure}
    \centering
    \includegraphics[width=\linewidth]{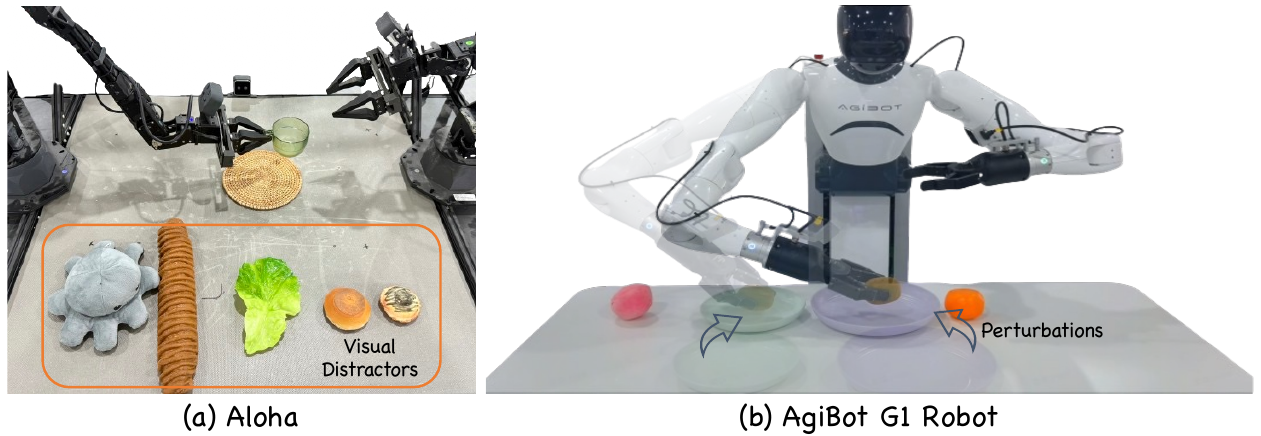}
    \caption{Hardware setup used in \Ours for both data collection and evaluation experiments. The Aloha robot is employed for conventional robotic policy experiments, which include various visual distractors. The AgiBot G1 robot is utilized for the VLA policy experiments, where different dynamic perturbations are applied.}
    \label{fig:hardware_setup}
\end{figure}

\subsection{\OURS in Conventional Robotic Policy}
\label{sec:c_policy}

To validate our proposed \OURS strategy, we first conduct experiments on conventional robotic policies trained with single-task imitation learning. For data collection and deployment, we use the widely adopted Aloha~\cite{zhao2023learning} hardware system and the ACT model as our baseline, as shown in Figure~\ref{fig:hardware_setup}(a). We evaluate the task \texttt{Pick the plastic cup and place it on the plate}, which requires the robot to grasp the cup's handle while addressing the challenge of transparent material. Our adversarial data collection strategy follow the concepts of \textit{dynamic visual perturbations} incorporating two key components: \textit{spatial diversity maximization} and \textit{dynamic target perturbation}. For a controlled comparison, we collect equal volumes of adversarially generated and conventional demonstration data.

Given the instability of the hardware and the fact that conventional robotic policy is not the primary focus of this paper, we provide only qualitative analysis. Policies trained on adversarially collected data demonstrate improved robustness and generalization performance. This improvement highlights the enhanced data density and state-action coverage in \OURS. However, we observe oscillatory grasping behavior when the placement height of the target object is varied. We attribute this to two main factors: (1) the inherent difficulty of depth estimation under increased state variance, and (2) the capacity limitations of smaller models cannot cover the high state coverage and low action consistency brought by \OURS, consistent with findings in~\cite{belkhale2023data}. These results provide empirical justification and motivation for scaling our methodology to VLA models.


\begin{table*}[t]
    \centering
    \renewcommand{\arraystretch}{1.1}
    \caption{Comparison of data collection strategies.}
    \begin{tabular}{c|c|c|c|c|c|c|c|c}
        \hline
        \rowcolor{gray!20} 
        \textbf{Method} & \textbf{Vis Perturb.} & \textbf{Lin. Perturb.} & \textbf{Varied Height} & \textbf{\#Epis.} & \textbf{\#Frame} & \textbf{Collection Time} & \textbf{Additional Label Time} & \textbf{Avg Time} \\
        \hline
        Traditional & $\times$ &  $\times$ & $\times$ & 120 & 90k & 25s per episode & 10s per episode & 46.7ms/frame \\
        \rowcolor{green!10} 
        \Ours   & \checkmark & \checkmark & \checkmark & 80 & 96k & 40s per episode & 15s per episode & 45.8ms/frame \\
        \hline
    \end{tabular}

    \label{tab:data_collection_comparison}
\end{table*}

\subsection{\OURS in VLA} 
\label{sec:method_in_vla}
To validate \OURS at the VLA scale, we implement our framework on the AgiBot G1 robotic platform using AR teleoperation for data collection~\cite{AgiBot2024agibotworld}, as shown in Figure~\ref{fig:hardware_setup}(b). We use $\pi_0$~\cite{black2024pi_0}, a state-of-the-art open-source VLA policy with demonstrated real-world manipulation capabilities, as our base model. Our evaluation focuses on the composite task \texttt{Grasp the [FRUIT-TYPE], place into the [CONTAINER]}, where \texttt{FRUIT-TYPE} $\in$ $\{$\texttt{orange, kiwi, orange}$\}$ and \texttt{CONTAINER} $\in$ $\{$ \texttt{green plate, blue plate} $\}$ . The pretrained model checkpoint~\cite{AgiBot2024agibotworld} is obtained through pretraining on the AgiBot-World dataset and is then fine-tuned on this task to enhance its performance.

This task requires both instruction grounding and visual target localization. To address these challenges, we design the following adversarial strategies:
\begin{itemize}
    \item Visual Perturbation. Dynamically displace target objects before grasping and reposition containers after pre-placement.
    \item Linguistic Perturbation. Modify instructions after successful grasping, e.g., changing "Place the orange into the green plate" to "Place the orange into the purple plate."
\end{itemize}

To manage annotation challenges from dynamic instructions, we employ task decomposition with subtask-level labeling (grasping/placement phases) while maintaining temporal continuity. Leveraging the VLA model's compositional generalization capability, we enable end-to-end policy learning from this structured input. Notably, we exclude kiwi placement from training data, leaving it as an unseen task for evaluation.

\section{Results and Analysis}

To evaluate \OURS for VLA models, we follow the strategy outlined in Section~\ref{sec:method_in_vla} and fine-tune the open-source $\pi_0$ model. To minimize modification effort, we adopt the default settings, using two wrist images and one head image as visual inputs. To assess the compositional generalization ability introduced by \OURS, we exclude the data for the task \texttt{Place the Kiwi into the [CONTAINER]} from the training set, expecting the model to generalize both action execution and language understanding to this unseen task. Each task is executed \textbf{10} times in the real world, and we report the success rate(SR). The data collected with different strategies for training the VLA model is summarized in Table~\ref{tab:data_collection_comparison}. The \texttt{Additional Label Time} is mainly attributed to language-visual alignment verification, as language instructions can vary for a single trajectory. Note that we define the \texttt{Episode} number as the scenario reset count, consistent with CALVIN~\cite{mees2022calvin}.

\subsection{Evaluating \OURS in Static Environments}
\label{sec:exp_in_static_env}
We begin with the evaluation in the most common static environment. To assess the model's performance more granularly, we report the SR for individual subtasks, such as \texttt{Pick} and \texttt{Place}, separately. The evaluation is conducted at three different worktable heights. Among these, \texttt{Var. 1} represents the standard table height, while \texttt{Var. 3} corresponds to an extreme height, pushing the robot to operate near its maximum reach limit. "normal positions" refer to cases where the target object is placed near the table's center, similar to the traditional data collection process, while "varied positions" involve sampling the target object's location across the entire work area. The evaluation results are summarized in Table~\ref{tab:evaluation_result_static}.
The VLA trained with \Ours outperforms the baseline VLA trained on the traditional-collection dataset across all heights and positions, achieving an average success rate of 0.72 compared to 0.13 at the most challenging height (\texttt{Var. 3}). It demonstrates strong compositional generalization on the \texttt{Place-C} task, while the model trained with traditional data  collection procedure struggles, particularly under varied positions (0.0). Notably, while both models perform well under the normal positions setting, the baseline model fails to handle varied positions, with a significant performance drop to 0.0 for all tasks.

\begin{table}[h]
    \centering
    \setlength{\tabcolsep}{3pt} 
    \renewcommand{\arraystretch}{1.1} 
    \arrayrulecolor{gray} 

        \caption{Evaluation results in static environment. The column \textit{Place-C} is highlighted for compositional generalization evaluation. }
    \begin{tabular}{c|c|ccc|ccc|c}
        \hline
        \hline
        \rowcolor{gray!20} 
        \textbf{Method} & \textbf{Height} & \multicolumn{3}{|c|}{\textbf{Normal Positions}} & \multicolumn{3}{|c|}{\textbf{Varied Positions}} & \textbf{Avg.} \\
        \rowcolor{gray!20} 
        & & Pick & Place & \cellcolor{yellow!40}Place-C & Pick & Place & \cellcolor{yellow!40}Place-C & \\
        \hline
        Traditional & Var. 1 & 1.0 & 0.8 & 1.0 & 0.0 & 0.0 & 0.0 & 0.47\\
        \Ours   &  & 1.0 & 1.0 & 1.0 & 1.0 & 1.0 & 1.0 & 1.0\\
        \hline
        Traditional & Var. 2 & 0.5 & 0.3 & 1.0 & 0.0 & 0.0 & 0.0 & 0.3\\
        \Ours   &  & 1.0 & 1.0 & 1.0 & 1.0& 1.0& 1.0 & 1.0 \\
        \hline
        Traditional & Var. 3 & 0.3 & 0.5 & 0.0 & 0.0 & 0.0  & 0.0 & 0.13\\
        \Ours   &  & 1.0 & 1.0 &  0.0 & 0.8 & 1.0 & 0.5 & 0.72\\
        \hline
        \hline
    \end{tabular}

    \label{tab:evaluation_result_static}
\end{table}

\subsection{Evaluating \OURS in Dynamic Environments}
\label{sec:exp_in_dynamic_env}
We further evaluate our method in dynamic environments. Specifically, we apply adversarial evaluation, similar to the approach used during data collection. For the visual aspect, a human evaluator modifies the positions of the target object or container before the effective actions are executed. For the linguistic aspect, the human evaluator alters the language instructions in real time. The evaluation experiments are conducted at a normal worktable height, with target objects placed across the entire workspace. The results are summarized in Tables~\ref{tab:evaluation_result_visual_dynamic} and~\ref{tab:evaluation_result_lingustic_dynamic}.
The results demonstrate that the model trained with \Ours exhibits significant robustness to visual perturbations. Specifically, dynamic changes in spatial positions have minimal impact on performance, achieving a SR of 0.88, comparable to the static environment evaluation. In contrast, the baseline model fails entirely under these conditions. For linguistic perturbations, we observe that performance is notably affected during the "During Grasp" phase. This is likely due to the fact that, once the gripper has already grasped or is very close to the object, changes in instructions require the policy to first halt the current action, re-evaluate the situation, and then decide on a new target, introducing significant challenges.

\begin{table}[h]
    \centering
    \setlength{\tabcolsep}{3pt} 
    \renewcommand{\arraystretch}{1.1} 
    \arrayrulecolor{gray} 

        \caption{Evaluation results in dynamic environment against the visual perturbations. }
    \begin{tabular}{c|ccc|ccc|c}
        \toprule
        \rowcolor{gray!20} 
        \textbf{Pert.} &  \multicolumn{3}{c|}{\textbf{Varied Container's Pos.}} & \multicolumn{3}{c|}{\textbf{Varied Object's Pos.}} & \textbf{Avg.} \\
        \rowcolor{gray!20} 
        \textbf{Method}& Pick & Place & Place-C & Pick & Place & Place-C & \\
        \midrule
        Traditional  & 0.0 & 0.0 & 0.0 & 0.0& 0.0& 0.0 & 0.0\\
        \Ours    & 0.8 & 0.7 & 1.0 & 0.8 & 1.0 & 1.0 &0.88\\

        \bottomrule
    \end{tabular}

    \label{tab:evaluation_result_visual_dynamic}
\end{table}

\begin{table}[ht]
    \centering
     \setlength{\tabcolsep}{3pt} 
    \renewcommand{\arraystretch}{1.1} 
    \arrayrulecolor{gray} 
        \caption{Evaluation results in dynamic environment against the linguistic perturbations.}
    \begin{tabular}{cc|cc|cc|cc}
        \toprule
        \rowcolor{gray!20} 
        \multicolumn{2}{c|}{\textbf{Pert. Time}} & \multicolumn{2}{c|}{\textbf{Before Grasp.}} & \multicolumn{2}{c|}{\textbf{During Grasp.}} & \multicolumn{2}{c}{\textbf{After Grasp.}} \\
        \cmidrule{1-8}
        \textbf{Methods} & & \textbf{Pick} & \textbf{Place} & \textbf{Pick} & \textbf{Place} & \textbf{Pick} & \textbf{Place} \\
        \midrule
        \multirow{1}{*}{Traditional} & & 0.0 & 0.0 & 0.0 & 0.0 & 0.0 & 0.0 \\
        \multirow{1}{*}{\Ours}   & & 1.0 & 1.0 & 0.6 & 0.7 & 1.0 & 1.0 \\
        \bottomrule
    \end{tabular}

    \label{tab:evaluation_result_lingustic_dynamic}
\end{table}

\subsection{Evaluating \OURS in Sensor-Failure Scenarios} 
\begin{figure}
    \centering
    \includegraphics[width=\linewidth]{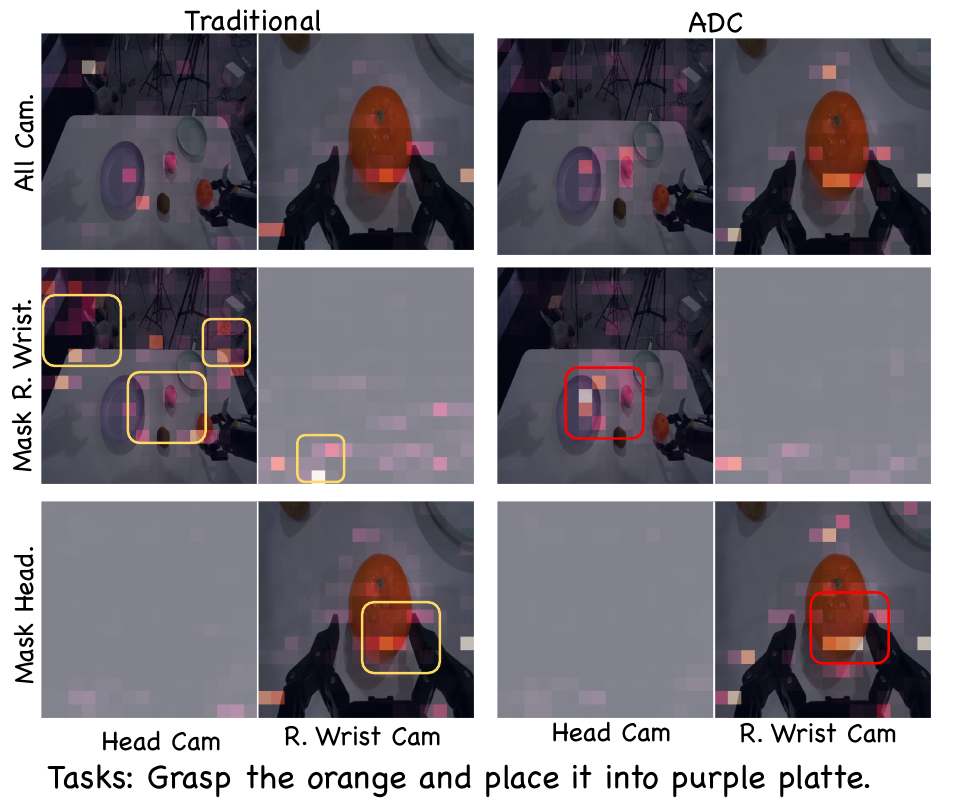}
    
    \caption{Comparison of attention maps when one camera is masked. Models trained with \Ours focus more precisely on functional cameras, demonstrating superior attention concentration compared to models trained with traditional data collection pipelines.}
    \label{fig:attention_map_masked_camera}
\end{figure}

We further evaluate the model trained with \Ours under extreme conditions, simulating scenarios where the equipped camera hardware fails. To replicate such failure scenarios, we replace the input from the target camera with an all-zero matrix before passing it to the VLA model. The evaluation results are presented in Table~\ref{tab:masked_camera_results}, demonstrating that the model trained with \Ours maintains consistent robustness across various masking settings.
To investigate the reasons behind this robustness, we visualize the attention maps between visual observations and action predictions extracted from the cross-attention layer. In this setup, the VLM's output tokens serve as inputs to the action expert header when a specific camera is masked. As shown in Figure~\ref{fig:attention_map_masked_camera}, the model trained with \Ours dynamically shifts its attention to functional cameras when a specific camera is masked. In contrast, the model trained with the traditional data collection process tends to focus more on irrelevant features, such as table edges, rather than the target objects when the wrist camera image is masked. The attention maps from the model trained with \Ours exhibit a more focused and meaningful response under masking conditions. Furthermore, \Ours data includes more instances of occlusion and scenarios where target objects are observed from multiple viewpoints, resulting in more comprehensive visual coverage, as illustrated in Figure~\ref{fig:observation_coverage_comparison}. This enriched data distribution significantly enhances the final policy model’s performance and robustness.

\begin{table}[ht]
    \centering
     \setlength{\tabcolsep}{3pt} 
    \renewcommand{\arraystretch}{1.1} 
    \arrayrulecolor{gray} 
    \caption{Evaluation results with masked cameras in static environment.}
    
    \begin{tabular}{cc|cc|cc|c}
        \toprule
        \rowcolor{gray!20} 
        \multicolumn{2}{c|}{\textbf{Masked Cam.}} & \multicolumn{2}{c|}{\textbf{Right Wrist}} & \multicolumn{2}{c|}{\textbf{ Head}} & \textbf{Avg.} \\
        \cmidrule{1-6}
        \textbf{Methods} & & \textbf{Pick} & \textbf{Place-AB} & \textbf{Pick} & \textbf{Place-AB} & \\
        \midrule
        Traditional  & & 0.0   & 0.0   & 0.0   & 0.0   & 0.0    \\
        \Ours    & & 0.6 & 0.5 & 0.7 & 0.4 & 0.55 \\
        \bottomrule
    \end{tabular}

    \label{tab:masked_camera_results}
\end{table}

\subsection{Further Analysis}
\label{sec:further_analysis}

\noindent \textbf{Data Efficiency.} As discussed in Section~\ref{sec:adversarial}, the data collected with \Ours demonstrates higher information density. This raises the hypothesis that significantly less \Ours data might be sufficient to train the VLA model while achieving comparable or even superior performance to a model trained on 100\% traditionally collected data. To test this hypothesis, we train additional VLA models under two conditions: (1) one model is trained with approximately 20\% \Ours data, and (2) another model is trained with 50\% \Ours data. These models are evaluated in static environments with both standard and varied object positions (as described in Section~\ref{sec:exp_in_static_env}) and in dynamic environments involving both visual and linguistic perturbations (as detailed in Section~\ref{sec:exp_in_dynamic_env}) on standard table height (\texttt{Var.1}). The results, summarized in Table~\ref{tab:performance_comparison}, reveal that even with only 20\% \Ours data, the model demonstrates significantly greater robustness and positional generalization in both static and dynamic environments compared to the model trained with 100\% traditionally collected data. Furthermore, as the proportion of \Ours data increases, the policy model's robustness improves further.

\begin{table}[ht]
    \centering
         \setlength{\tabcolsep}{4pt} 
    \renewcommand{\arraystretch}{1.1} 
    \arrayrulecolor{gray} 
    \caption{Performance comparison of different datasets in static and dynamic environments.}
    \label{tab:performance_comparison}
    \begin{tabular}{lcc|cc|c}
        \toprule
        
         & \multicolumn{2}{c}{\textbf{Static Env.}} & \multicolumn{2}{c}{\textbf{Dynamic Env.}} &  \\
        \cmidrule(lr){2-3} \cmidrule(lr){4-5} 
         \textbf{Dataset Receipt} & \textbf{Pick} & \textbf{Place} & \textbf{Pick} & \textbf{Place} & \textbf{Avg.} \\
        \midrule
        100\% Traditional & 0.5  & 0.45  & 0.0  & 0.0   & 0.24 \\
        20\% Ours    & 0.5  & 0.75  & 0.58 & 0.75  & 0.65 \\
        50\% Ours    & 0.83 & 0.75  & 0.63 & 0.75  & 0.74 \\
        100\% Ours   & 0.9  & 0.875 & 0.83 & 0.94  & 0.89 \\
        \bottomrule
    \end{tabular}
\end{table}

    

\begin{figure}[h]
    \centering
    \includegraphics[width=\linewidth]{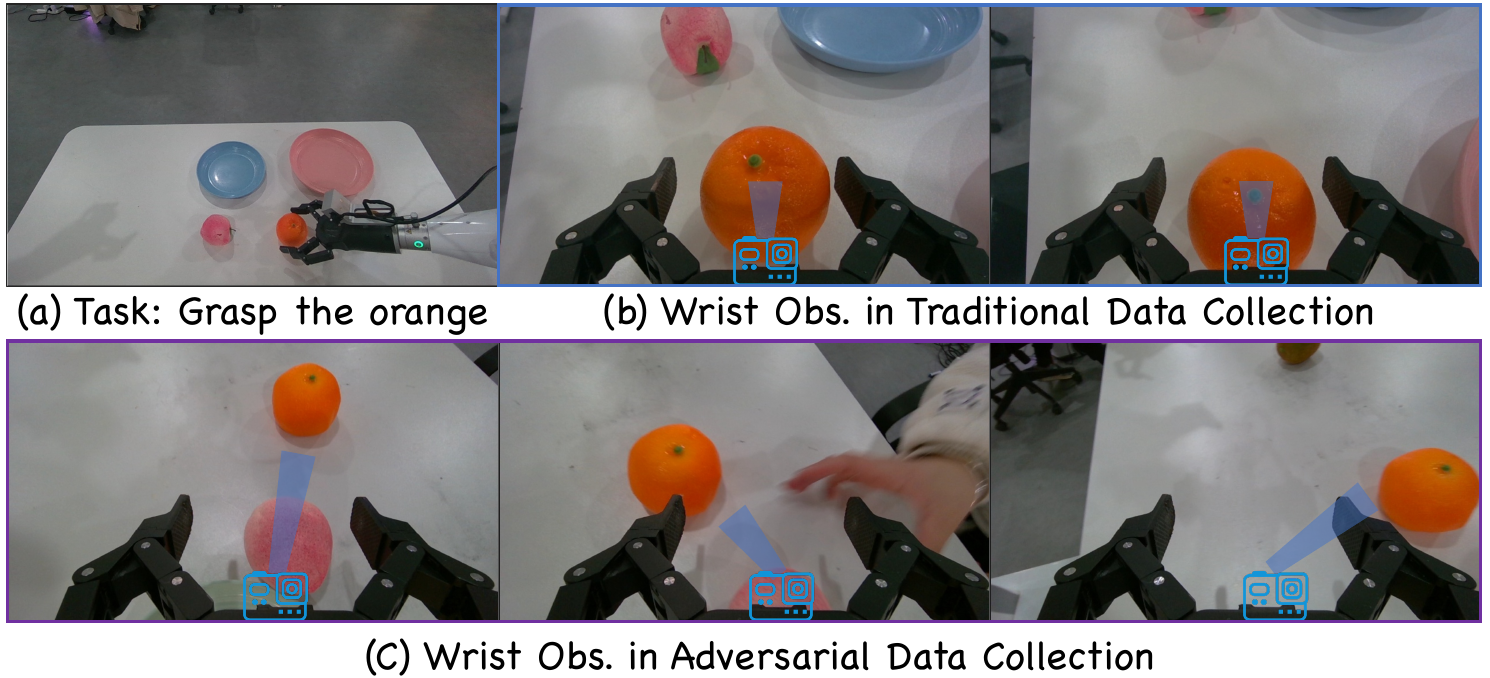}
    \caption{Comparison of observation coverage for the task "Grasp the orange." In the traditional data collection process, the target object (orange) is observed from similar viewpoints, resulting in limited visual diversity. In contrast, \Ours introduces dynamic perturbations, allowing the orange to be observed from a wider range of viewpoints. This leads to greater visual variation in the \Ours dataset, improving model robustness and generalization.}
    \label{fig:observation_coverage_comparison}
\end{figure}

\noindent \textbf{Complete Object Observation Coverage.} We analyze the benefits of \OURS by examining the visual observation coverage for the task Grasp the orange. As shown in Figure~\ref{fig:observation_coverage_comparison}, compared to wrist observation images collected using the standard strategy, those collected with \Ours exhibit substantially greater variation in the orange's visual appearance. This increased diversity in observation enhances the model’s generalization capabilities while reducing the need for extensive data collection.

\begin{figure}
    \centering
    \includegraphics[width=\linewidth]{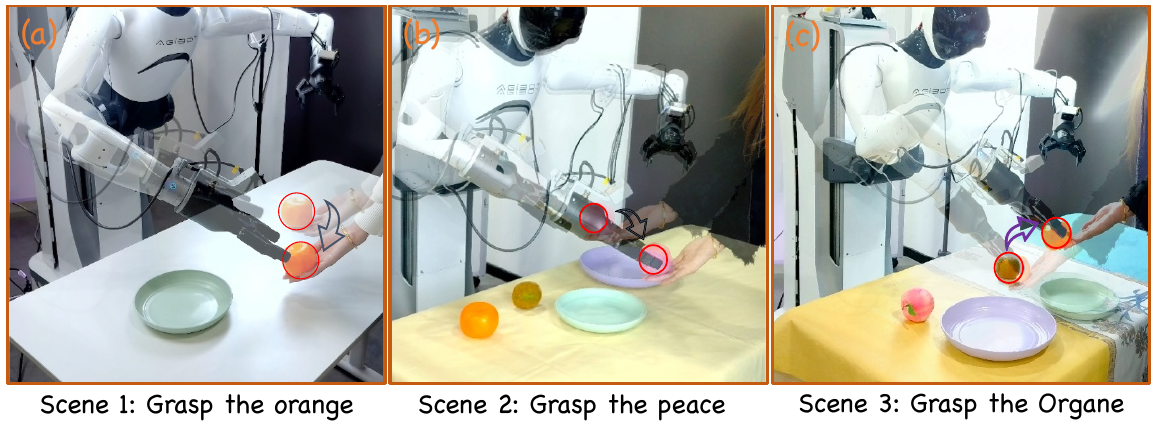}
    
    \caption{Dynamic Human-Robot Interaction (HRI) scenarios. The robot is tasked with grasping the target fruit from the human hand, where the human’s hand may move during the manipulation tasks. Evaluation experiments are conducted across different scenes. }
    
    \label{fig:dynamic_hri}
\end{figure}

\noindent \textbf{Dynamic HRI.} Previous deployments and evaluations of VLA models have primarily focused on static environments, either in the real world or in simulators. However, the ultimate goal of robotics is seamless integration with humans, requiring robots to provide dynamic and adaptive responses during task execution, even under human intervention. To evaluate this capability, we design two task variations: (1) a human holds and moves the target object during grasping, altering its pose and height; and (2) language perturbation, as described in Section~\ref{sec:exp_in_dynamic_env}. These scenarios require the robot to dynamically adjust its action predictions. The experimental setup is illustrated in Figure~\ref{fig:dynamic_hri}. Models trained with \OURS demonstrate the ability for dynamic adaptation, a critical prerequisite for future human-robot interaction (HRI).

\noindent \textbf{Shuffle Buffer.} As discussed in \cite{team2024octo}, effective shuffling of frames during training is crucial for the final performance of VLA models, ensuring that each batch contains diverse information. Their approach required significant engineering effort to interleave images from different trajectories. In contrast, trajectory data collected via \OURS inherently includes diverse motion and semantic information even within a single trajectory. This significantly reduces the engineering effort needed to manage large shuffle buffers.

\begin{figure}
    \centering
    \includegraphics[width=\linewidth]{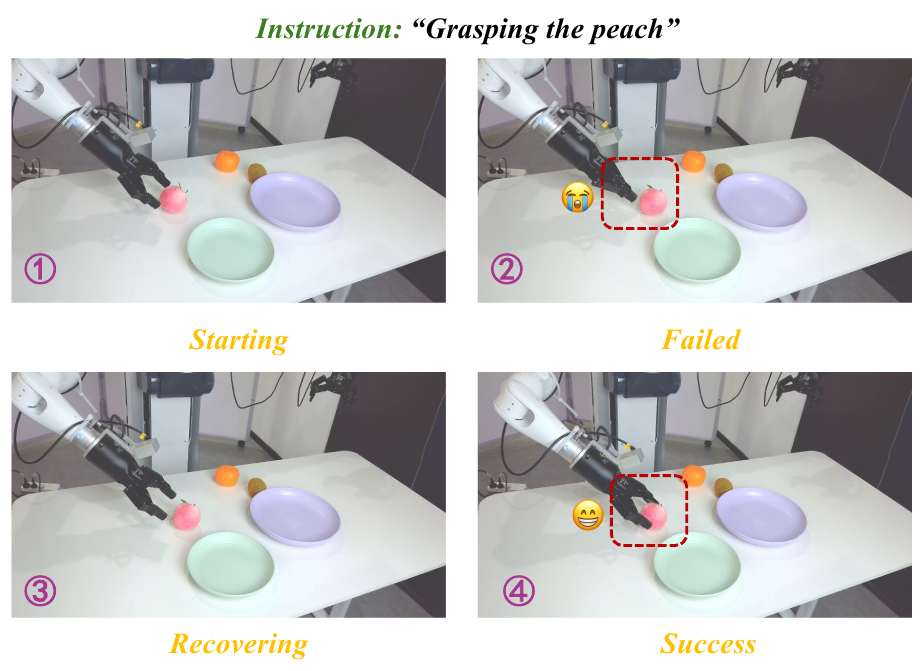}
    
    \caption{Autonomous Failure Recovery in ADC-Trained Robotic Grasping: Real-time demonstration of failure recovery after empty grasp. Following initial contact loss during peach acquisition, the system autonomously recalibrates grip pose parameters and executes a precision-aligned regrasp to complete the task.}
    
    \label{fig:fail_recover}
\end{figure}

\noindent \textbf{Data Collection Cost Discussion.} While both \OURS and conventional methods exhibit comparable per-frame annotation latency (45.8ms vs. 46.7ms) as shown in Table~\ref{tab:data_collection_comparison}, their human cost profiles differ substantially. \OURS requires two operators: a tele-operator for task execution and an adversarial operator for perturbations. Though doubling the labor count, the adversarial operator demands significantly lower expertise and intermittent involvement compared to continuous tele-operation.  Critically, \OURS’s marginally higher episodic time cost ($<2$×) is counterbalanced by its $>5$× improvement in data efficiency. This tradeoff—minimal additional labor for exponential gains in sample quality—establishes ADC as a practical solution for scalable robotic learning in cost-constrained settings.

\noindent \textbf{Novel Scene Generalization.} Although the \Ours data is collected without tablecloths, using only the original white table surface, the final VLA policy demonstrates strong generalization capabilities in unseen, novel scenes. As shown in Figure~\ref{fig:dynamic_hri}, evaluation experiments are conducted in different scenes where tablecloths are introduced. We attribute this generalization ability to two key factors: (1) The strong pre-trained visual encoders employed in the VLA model. (2) The \Ours data collection procedure, which includes more visually adversarial images, making the trained model robust and generalizable. In contrast, the same model architecture trained with traditional data performs worse in zero-shot unseen scenarios.

\noindent\textbf{Failure Recovery Analysis.}The VLA model trained on our ADC dataset demonstrates an autonomous failure recovery ability. As shown in Figure~\ref{fig:fail_recover}, the robotic arm automatically initiates a secondary grasping attempt upon detecting an empty-gripper state. This capability is enabled by the enhanced robustness gained through systematically injected disturbances during \Ours-based training. Two key factors brought by \Ours contribute to this failure recovery ability: (1)The \Ours data includes diverse grasp initialization poses, allowing the learned policy to generalize to a broader range of scenarios. (2) During the adversarial data collection process, tele-operator failures are inevitable due to the injected disturbances. When failures occur, the tele-operator adjusts and reattempts the task. This retry data further enhances the model's ability to recover from failure.

\section{Conclusion}
In this paper, we demonstrate that strategic adversity during data acquisition is fundamental to efficient robotic learning. Our Adversarial Data Collection (ADC) framework proves that human-driven perturbations in visual and linguistic dimensions can compress hundreds of static variations into single demonstrations, directly resolving the data redundancy and domain gaps inherent to conventional pipelines. By integrating real-time human adaptation with physics-constrained perturbations, ADC-trained models inherently acquire robustness to environmental uncertainties and compositional language variations—capabilities unattainable through scale-driven data collection. This work establishes intentional adversity as a new paradigm for robotic learning, where data quality, orchestrated through human-environment interplay, supersedes brute-force quantity. These insights compel a re-examination of robotic data ecosystems: purposeful perturbation during acquisition, not just algorithmic innovation, may hold the key to generalizable embodied intelligence.

\bibliographystyle{IEEEtran}
\bibliography{IEEEabrv,iros}



\end{document}